\documentclass[conference]{IEEEtran}
\IEEEoverridecommandlockouts
\usepackage{amsmath,amssymb,amsfonts}
\usepackage{algorithmic}
\usepackage{graphicx}
\usepackage{textcomp}
\usepackage[table]{xcolor}

\usepackage[sort]{natbib}
\usepackage{subfigure}
\usepackage{times}
\usepackage{makecell}
\usepackage{multirow}
\usepackage{float}
\usepackage{titlesec}

\begin{document}
\title{Norm-Q: Effective Compression Method for Hidden Markov Models in Neuro-Symbolic Applications}


\author{
\IEEEauthorblockN{Hanyuan Gao \quad Xiaoxuan Yang}
\IEEEauthorblockA{
\textit{Department of Electrical and Computer Engineering} \\
\textit{University of Virginia}\\
Charlottesville, USA \\
\{zqv4ds, xiaoxuan\}@virginia.edu} \\
\vspace{-1.2cm}
}

\maketitle

\begin{abstract}
Hidden Markov models (HMM) are commonly used in generation tasks and have demonstrated strong capabilities in neuro-symbolic applications for the Markov property. These applications leverage the strengths of neural networks and symbolic reasoning to create robust and interpretable AI systems. However, they may inherit and amplify the shortcomings of both approaches. Both components require dense computation and data transfer, and their communication further hinders performance. This paper proposes Norm-Q, a normalized linear quantization approach for compressing probabilistic symbolic models, such as HMMs. We reduce the bit width of the data with minimal impact, thereby alleviating memory and bandwidth stress and enabling deployment on potential custom hardware. Our method introduces a normalized quantization-aware expectation maximization process for probabilistic model training. The experimental results show that Norm-Q achieves a higher compression rate with reasonable score loss compared to traditional quantization methods. In the case of the constrained generation task of large language models, we successfully quantize an HMM of 4096 hidden states to 8 bits without loss and, at most, 3 bits with acceptable loss. Notably, the Norm-Q method can achieve a compression rate of $99\%$ for the weights of the HMM. The code is open source at https://github.com/superstarghy/Norm-Q.
\end{abstract}


\section{Introduction}

Neuro-symbolic models are emerging as the next generation of machine learning model architecture \cite{susskind2021neuro}. This architecture combines the data processing capacities of neural networks with the reasoning capabilities of symbolic AI, resulting in reasonable, interpretable, and robust AI systems \cite{garcez2023neurosymbolic}. Meanwhile, the integration of these computing paradigms presents significant challenges, particularly in efficiently supporting both paradigms within the system \cite{wan2024towards, yang2023neuro}. Data transfer requirements between these two parts can severely hinder performance. For example, on the constrained generation task of large language models (LLM) \cite{lin2019commongen}, the latency overhead increased by 4x when a symbolic model was integrated \cite{zhang2025adaptable}. The response time prohibits their deployments as real-time applications. 

\begin{figure}[t]
    \centering
    \resizebox{\columnwidth}{!}{
    \hspace{-1em}
    \subfigure[Neural part]{
        \includegraphics[width=0.45\columnwidth]{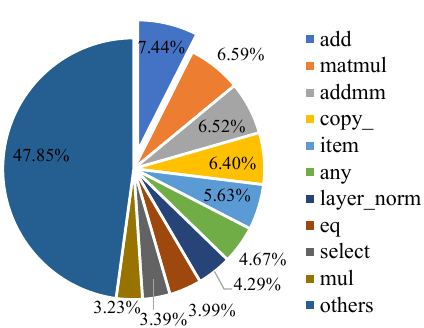}
        \label{fig:1.1}
    }
    \hspace{-1em}
    \subfigure[Symbolic part]{
        \includegraphics[width=0.35\columnwidth]{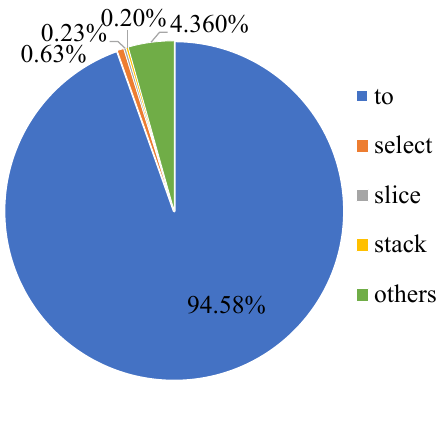}
        \label{fig:1.2}
    }
    \hspace{-1em}
    \subfigure[Latency results]{
        \includegraphics[width=0.32\columnwidth]{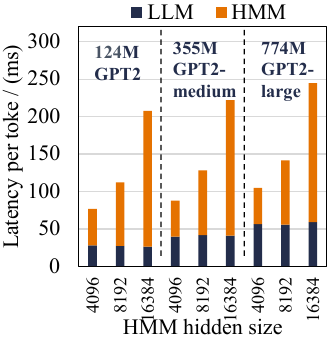}
        \label{fig:1.3}
    }
    }
    \vspace{-10pt}
    \caption{Latency profiling of an LLM integrated with an HMM.}
    \label{fig:1}
    \vspace{-15pt}
\end{figure}

From a system perspective, understanding the different and conflicting demands of the two components is crucial to analyzing the severe latency issue.
On one hand, neural models, consisting of matrix multiplications (MatMul) and convolutions, are computation-intensive.
On the other hand, symbolic models are typically memory-intensive models, where vectors or element-wise operations have a larger proportion.
Furthermore, when combining two counterparts, the extra communication emerges as a bottleneck due to the different nature of the operations. 
Fig.~\ref{fig:1.1} and \ref{fig:1.2} show the PyTorch operator profile result of a GPT2-large model, the neural part, combined with a hidden Markov model (HMM) and an automata, the symbolic part \cite[]{zhang2025adaptable}.
In the symbolic part, memory copy and data transfer operations account for over $95\%$.
Moreover, we find that the scaling effects are different for the neural and symbolic parts. Fig.~\ref{fig:1.3} shows the latency variation across various model scales. HMM has a more significant impact on performance than LLM. Specifically, LLM latency increases by 1.45x when its size doubles, while the HMM corresponding latency scale factor is up to 2x. Therefore, optimization and compression of symbolic models are necessary to ensure the overall efficiency of neuro-symbolic applications.

Traditional optimization methods for the neural part, e.g., pruning \cite[]{fang2023depgraph}, integer quantization \cite[]{gholami2022survey}, and clustering \cite[]{han2015deep, itay2018qnn}, cannot be directly applied to the symbolic part due to two concerns.
First, in probabilistic models, weights carry specific semantic meanings that are crucial to their functionality. For example, tractable probabilistic models leverage the probability distribution to provide the logits for tokens generated from LLM. Therefore, directly applying conventional optimization methods to these weights can severely disrupt the data distribution characteristics.
Second, as the dimension of distribution increases, the probability values after quantization tend to decrease to zero.
For example, quantized weight matrices with an HMM hidden size of $16384$ are more likely to contain rows of all zero values compared to matrices with a smaller HMM hidden size of $4096$. For HMM with a large hidden size, quantization may result in rows containing exclusively zero values due to rounding effects. In such circumstances, the matrix will contain erroneous information, which may result in meaningless or incorrect output for the generation stage. Existing concerns need to be addressed with a new quantization method compatible with probability models.

This paper targets tractable probabilistic models and aims to alleviate memory and bandwidth stress, tackling the challenges of neuro-symbolic model acceleration.
We focus on the acceleration of representative HMM, a large-scale model commonly used in symbolic parts.
A row normalization quantization method (Norm-Q) is proposed to address the inaccuracy and decay to zero issues. Norm-Q enables outstanding constraint success rates and generation quality scores compared with prior methods.
Furthermore, we introduce a row-normalized linear quantization-aware expectation maximization (Norm-Q aware EM) process for probabilistic model training. 
The compression of symbolic components, which is orthogonal to the optimization of neural parts, is the main emphasis of this work.
The experimental results show that our Norm-Q method demonstrates 8-bit quantization without score loss for the HMM of 4096 hidden states, which outperforms the existing quantization and pruning methods. In contrast, the previous integer method exhibits an unacceptable success rate degradation of more than $70\%$ at 8-bit precision. When further quantized to 4 bits, our method shows a minimal success rate reduction of less than $1\%$, and the final score degradation of generation is under $2\%$.
In addition, our method demonstrates no deterioration when HMMs scale up. The success rate and score loss for an HMM with a hidden size of 16,384 are under $4\%$.
Overall, the HMM weights are compressed by more than $99\%$ with the effective Norm-Q method.

\section{Preliminaries}
\label{sec:pre}
This section introduces the basis for neuro-symbolic models, hidden Markov models, and the quantization methods. We then discuss related work on HMM compression.

\noindent \textbf{Neuro-symbolic Model.} \quad The model combines neural models, e.g., a convolutional neural network or LLM, with symbolic models. Symbolic models consist of logic operations \cite[]{dong2019neural, hsu2024s} and probability calculation \cite[]{ahmed2024controllable, han2019visual, zhang2021abstract, choi2020probabilistic}, providing reasoning capability for the application. Neuro-symbolic models leverage the data featuring capability from neural networks and reasoning ability from logic operations to promote AI performance.
Another advantage of using a symbolic part is that the model adopts knowledge and rules such that the data for training the neural part is greatly reduced \cite[]{badreddine2022logic}. Unfortunately, the extra latency overhead is huge after introducing the symbolic part, especially the data transmission between different parts exacerbates the bandwidth demand.

\noindent \textbf{Hidden Markov Model.} \quad A hidden Markov model is a Markov model in which the observations depend on states, and the states are latent. HMMs are commonly used in generation tasks, e.g., LLM \cite[]{zhang2025adaptable} and diffusion \cite[]{ho2020denoising}. The Markov property derives HMM's reasoning capacity \cite[]{rabiner1986introduction} in tasks such as conceptual generation \cite[]{lin2019commongen, zhang2023tractable}.
. The initial probabilities matrix $\gamma$ or $P(z_0)$,  transition matrix $\alpha$ or $P(z_{t+1}|z_t)$, and emission matrix $\beta$ or $P(x_t|z_t)$, define an HMM. The sizes of the three matrices are [1, hidden size], [hidden size, hidden size], and [hidden size, output size], respectively. The following equation shows the forward algorithm \cite[]{rabiner1986introduction}, where $x_t$ and $z_t$ denote the output and state at $t$:
\[
    P(x_{1...t}, z_{t+1}) = \sum_{Z_t} P(z_t, x_{<t})P(x_t|z_t)P(z_{t+1}|z_t).
\]

\noindent \textbf{Quantization and Pruning} \quad Neural network quantization \cite[]{liang2021pruning} utilizes the robustness of neural networks under different data precisions and reduces the bit width or representation space of data. The method uses a cookbook to represent all data points. The following function $Q$ shows the theory of quantization, which quantizes 32-bit floating point (FP32) format data to a $b$ bits indexed cookbook as follows: 
\begin{align*}
  \{p_1, p_2, ..., p_n\} & \overset{\text{Q}}{\mapsto} \{q_1, q_2, ..., q_n\}, \ p_i \in R_{FP32}, \\
  & q_i \in Cookbook, |Cookbook|= 2^b \ll |R_{FP32}|,
\end{align*}
\noindent
where $p$ denotes the values before quantization, $q$ denotes the quantized values, $R$ denotes the range, and $Cookbook$ indicates a lookup table or predefined set of values used for mapping continuous float point data to a more compact representation.
Eventually, each value is represented as an index. The inverse process is called dequantization. The mapping is implemented by various quantization algorithms, such as uniform quantization and clustering \cite[]{gholami2022survey}, which are broadly categorized into linear and non-linear methods. Linear quantization utilizes a scale factor and a zero point to transform numbers into integers:
\[
    q_i = clip(round( p_i \times scalefactor ) + zeropoint ). 
\]
Pruning is another commonly used method to compress a model, which directly eliminates zeros and small values, leveraging model sparsity.

\noindent \textbf{Compression for HMM} \quad There is limited research on HMM data compression. Prior work demonstrates the use of HMMs as tools for compressing other models or data types rather than compressing the HMMs themselves~\cite[]{li2000image, han2024prediction}. Wu et al. focused on reducing the number of hidden states, which affects the model structure rather than parameter representation~\cite{wu2010probability}. Bello et al. proposed optimization from an algorithmic perspective, enhancing the efficiency of forward and backward calculations of HMMs~\cite{bello2023compressed}. Notably, these existing approaches lack a system-aware methodology, considering the holistic impact of compression on the entire neuro-symbolic system. Moreover, a parameter size reduction method dedicated to HMM is worth exploring. 

\section{Methodology}
Our conceptual generation task requires generating sentences with target concepts or keywords. Since the current output is generated based on the previously generated tokens, it is more difficult to ensure that future tokens satisfy the target concepts or follow a specific keyword sequence.
This is where HMMs come into play, helping manage and enforce these constraints. Moreover, the task not only needs to meet the constraints but also requires generation quality, which is evaluated by the constraint success rate and sentence quality scores. This section presents the techniques and considerations for compressing HMM weights. 

We explore various approaches to identify the most effective one since it is the first attempt to quantize an HMM, . 
To address the limitations of existing approaches, we propose Norm-Q, a normalized quantization method specifically for HMMs.
The Norm-Q method provides balanced and outstanding success rates and generation quality scores. Furthermore, we propose the Norm-Q Aware Expectation Maximization (Norm-Q aware EM) to enable efficient HMM training.



\begin{figure}[t]
    \vspace{-5pt}
    \small
    \resizebox{\columnwidth}{!}{
    \hspace{-2em}
    \subfigure[Heat map of $\alpha$ matrix]{
        \centering
        \includegraphics[width=0.47\columnwidth]{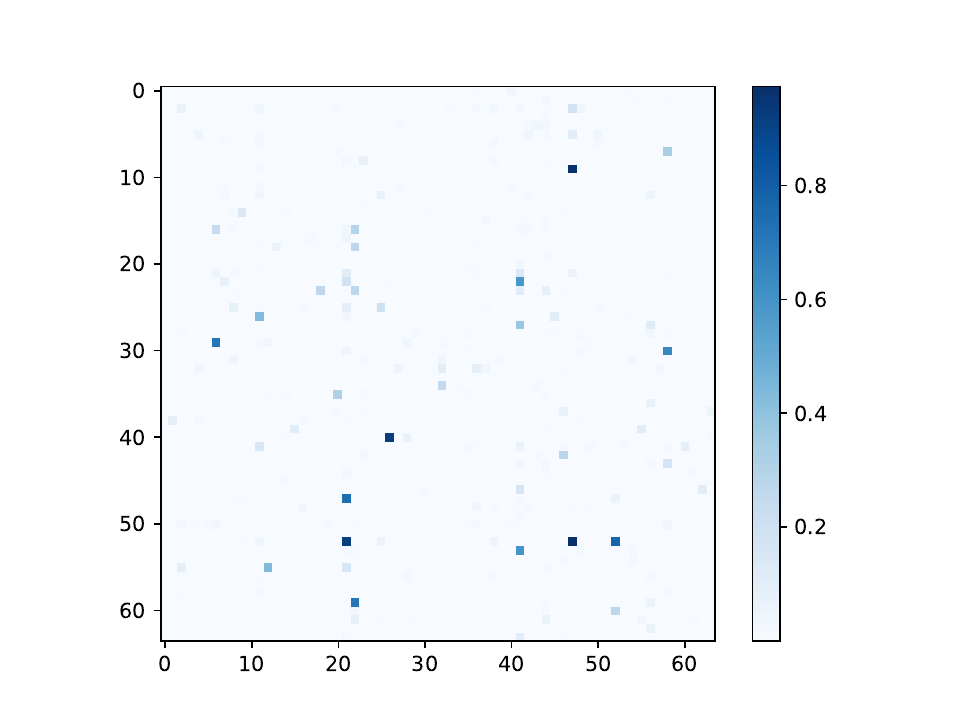}
        \label{fig:2.1}
    }
    \hspace{-2em}
    \subfigure[Distribution of $\beta$ matrix]{
        \centering
        \includegraphics[width=0.52\columnwidth]{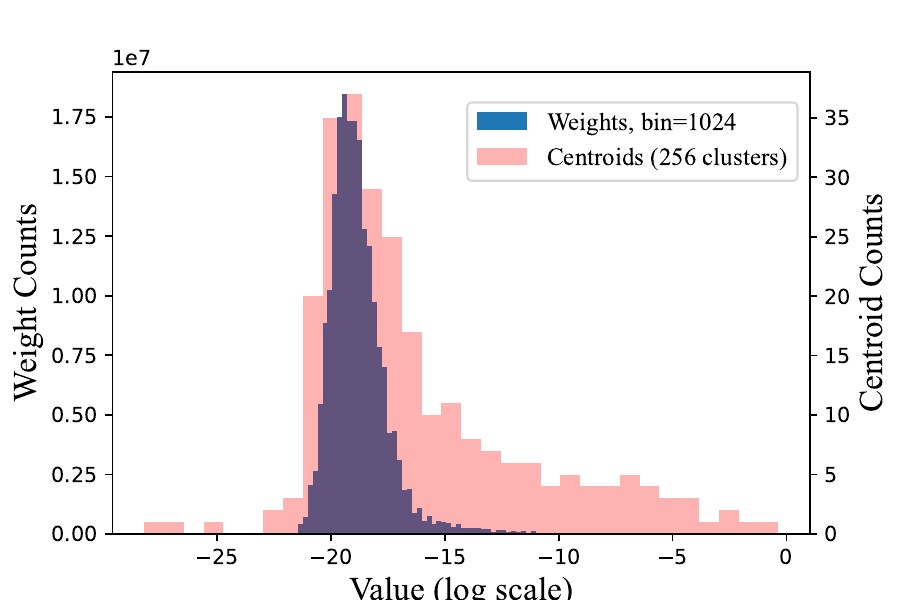}
        \label{fig:2.2}
    }
    }
    \caption{Data distribution of transition ($\alpha$) and emission ($\beta$) matrices of an HMM. The heat map is obtained by max-pooling and sampling to 64x64.}
    \label{fig:2}
    \vspace{-10pt}
\end{figure}

\subsection{HMM Weight Distribution Analysis and Pruning}


We utilize an HMM with an output dimension of 50257 and a hidden dimension of 4096 for case study. Note that HMM can be generated via LLM distillation or trained from scratch. More details of experimental settings can be found in Section~\ref{section:experiments}. 
As introduced in Section~\ref{sec:pre}, HMM has three weight matrices. Since the initial matrix ($\gamma$) has the smallest number of parameters, we analyze the data distributions of transition ($\alpha$) and emission ($\beta$) matrices in Fig.~\ref{fig:2}. In both matrices, element values smaller than $10^{-5}$ account for more than $80\%$. This large portion of small values within HMM weight matrices motivates us to explore pruning first. However, since the matrices within probabilistic models carry semantic information, 
the pruning changes the values and may disturb the final output. Therefore, there is an inherent trade-off between the sparsity ratio and HMM model effectiveness.

Our experiments prune the HMM weights to a certain ratio, followed by the generation task in the neuro-symbolic application. The constraint success rate and score results are shown in Table~\ref{tab:pruning}.  When pruning ratio is set as $85\%$, the outputs remain mostly unaffected, i.e., the rate and scores change within $0.2\%$. However, when the pruning ratio reaches $86\%$, the model generates wrong and garbled outputs. Therefore, the constraint success rate will drop to 0, and the evaluation scores are invalid.  The result illustrates that this HMM tolerates a certain amount of pruning, and the model’s functionality can be compromised beyond a certain level of pruning.

\begin{table}[t]
    \centering
    \small
    \caption{Constraint success rate and score (x100\%) results of ratio-based pruning}
    \vspace{-5pt}
    \resizebox{0.9\columnwidth}{!}{
    \begin{tabular}{c|c|c|c|c|c|c}
        \hline
        Pruning Ratio & 50\% & 80\% & 85\% & 86\% & 90\% & 86\% w/ norm \\
        \hline
        Sucess Rate & 99.8 & 100 & 100 & 0 & 0 & 82.0 \\
        Rouge & 37.6 & 37.5 & 37.4 & - & - & 37.4 \\
        BLEU4 & 35.1 & 34.7 & 34.5 & - & - & 34.5 \\
        CIDER & 11.5 & 11.4 & 11.4 & - & - & 11.4 \\
        SPICE & 26.9 & 27.0 & 27.0 & - & - & 27.0 \\
        \hline
    \end{tabular}
    }
    \label{tab:pruning}
    \vspace{-5pt}
\end{table}
At higher pruning ratios,  all elements in certain matrix rows are pruned to zero. These rows represent probability distributions, and when pruned entirely, the information within these rows is lost and cannot be recovered. For example, a row of the emission matrix corresponds to the probability of a state generating different tokens. If all probabilities in a row are zero, the state can no longer generate any tokens. Therefore, ratio-based pruning alone cannot fully exploit the compression potential of the HMM, as it suffers from instability and information loss, particularly when entire rows are pruned. 

\subsection{Layer-wise Quantization Approach}

In addition to reducing the number of weights, reducing the storage space occupied by a single weight further compresses the model.
Quantization will force values to be approximated to certain points, destroying the distribution. Moreover, the characterization ability of the model will decrease due to the reduction of data diversity.
To mitigating these effects, we target at HMMs that cooperate in neural symbolic applications, leveraging the robustness of the neural part.

We first conduct layer-wise quantization that transforms the values before an operation and recovers them afterward. Layer-wise quantization requires de-quantization such that the transformation function must be reversible during computations, such as a multiply-add operation:

\scalebox{0.9}{
\begin{minipage}{\columnwidth}
\begin{align*}
    DQ(q_1 \times q_2 + q_3 \times & q_4) = DQ[Q(p_1) \times Q(p_2) + Q(p_3)\times Q(p_4)] \\
    &\approx p_1 \times p_2 + p_3 \times p_4,
\end{align*}
\end{minipage}
}\\

\noindent $Q$ represents the quantization process and $DQ$ represents de-quantization. $DQ$ is approx to $Q^{-1}$ in general. Non-linear quantization functions cannot be used in the layer-wise method because they do not satisfy the above requirement.
We experiment with linear layer-wise quantization, where de-quantization is implemented by dividing a factor. We quantize and dequantize the numbers during four main MatMul layers in the application \cite[]{zhang2025adaptable}.
The results of different bit widths in Table~\ref{tab:int} show that the success rate drops by over $10\%$ below 12 bits.
Then, we tried quantizing all weights to integer numbers, and the generation results became garbled and incorrect. Thus, previous integer-based quantization methods designed for neural networks cannot be applied to probabilistic models for low-precision compression.

\begin{table}[t]
    \centering
    \small
    \caption{Constraint success rate and score (x$100\%$) results of layer-wise integer quantization}
    \vspace{-5pt}
    \resizebox{\columnwidth}{!}{
    \begin{tabular}{
    c|c
    |@{\hspace{2pt}}c@{\hspace{1pt}}|@{\hspace{2pt}}c@{\hspace{2pt}}|@{\hspace{2pt}}c@{\hspace{2pt}}|@{\hspace{2pt}}c@{\hspace{2pt}}
    |@{\hspace{2pt}}c@{\hspace{2pt}}|@{\hspace{2pt}}c@{\hspace{2pt}}|@{\hspace{2pt}}c@{\hspace{2pt}}|@{\hspace{2pt}}c@{\hspace{2pt}}}
        \hline
        Bitwidth & \cellcolor{gray!30}FP32 & INT24 & INT16 & INT14 & INT12 & INT11 & INT10 & INT9 & INT8 \\
        \hline
        Success Rate & \cellcolor{gray!30}100 & 100 & 99.5 & 97.8 & 89.0 & 78.0 & 56.0 & 34.0 & 21.0 \\
        Rouge & \cellcolor{gray!30}37.7 & 37.8 & 37.8 & 37.8 & 37.4 & 36.7 & 34.8 & 31.8 & 29.1 \\
        BLEU4 & \cellcolor{gray!30}34.4 & 34.4 & 34.8 & 35.1 & 35.2 & 34.8 & 34.0 & 31.9 & 28.8 \\
        CIDER & \cellcolor{gray!30}11.7 & 11.7 & 11.7 & 11.7 & 11.6 & 11.2 & 10.1 & 8.26 & 6.65 \\
        SPICE & \cellcolor{gray!30}27.8 & 27.8 & 27.7 & 27.7 & 27.0 & 25.6 & 22.9 & 19.1 & 16.0 \\
        \hline
    \end{tabular}
    }
    \label{tab:int}
    \vspace{-10pt}
\end{table}
For a probabilistic distribution, the quantization is a constrained optimization problem and a clustering process that keeps the distribution before and after quantization as close as possible. KL-divergence is a proper loss function to measure the difference between two distributions, as shown below.
\scalebox{0.85}{
\begin{minipage}{\columnwidth}
\begin{align*}
   \hat{P} = &\{p_1, ..., p_n\}, \hat{Q} =\{q_1, ..., q_n\},
   q_i \in Cookbook, \ \ |Cookbook|= 2^b \\
   & argmin_{\hat{Q}} loss(\hat{P},\hat{Q}) = D_{KL}(\hat{P}||\hat{Q})
   \Leftrightarrow NLLloss(\hat{Q},\hat{P}).
\end{align*}
\end{minipage}
}\\

\noindent The divergence is equivalent to negative log-likelihood loss (NLLloss). If $Q$ is not constrained, the solution will be $Q=P$ according to Gibbs' inequality.
K-means clustering can be a non-linear and iterative solution to the problem, but the results are not as expected, as shown in the Direct K-means row of Table~\ref{tab:kmeans}. K-means uses floating-point centroids as a cookbook and clusters all numbers, one-dimensional in our context, to these centroids. 
When clustering to 256 centroids, i.e., 8 bits, the success rate drops by $36\%$. The clustering approach outperforms the integer-based method on INT8 by $40\%$ in rate and $6\%$ in scores on average. Therefore, the K-means method cannot ensure the constraint requirement.
\begin{table}[t]
    \centering
    \caption{Results (x100\%) of 256 centroids K-means}
    \vspace{-5pt}
    \resizebox{0.85\columnwidth}{!}{
    \begin{tabular}{@{}c@{\hspace{2pt}}|@{\hspace{2pt}}c@{\hspace{2pt}}|@{\hspace{2pt}}c@{\hspace{2pt}}|@{\hspace{2pt}}c@{\hspace{2pt}}|@{\hspace{2pt}}c@{\hspace{2pt}}|@{\hspace{2pt}}c@{\hspace{2pt}}}
        \hline
            & Success Rate & Rouge & BLEU4 & CIDER & SPICE \\
        \hline
        Direct K-means & 64.0 & 35.8 & 33.9 & 10.8 & 24.6 \\
        K-means during EM & 81.0 & 36.9 & 35.0 & 11.6 & 25.2 \\
        \hline
    \end{tabular}
    }
    \label{tab:kmeans}
    \vspace{-10pt}
\end{table}

\subsection{Fixed-point Linear Quantization}

\begin{table}[t]
    \centering
    \small
    \caption{Sparsity (zero ratio, x100\%) after auto-pruning of fixed-point linear quantization}
    \vspace{-5pt}
    \resizebox{\columnwidth}{!}{
    \begin{tabular}{
    @{\hspace{2pt}}c@{\hspace{2pt}}|@{\hspace{2pt}}c@{\hspace{2pt}}|@{\hspace{2pt}}c@{\hspace{2pt}}|@{\hspace{2pt}}c@{\hspace{2pt}}|@{\hspace{2pt}}c@{\hspace{2pt}}
    |@{\hspace{2pt}}c@{\hspace{2pt}}|@{\hspace{2pt}}c@{\hspace{2pt}}|@{\hspace{2pt}}c@{\hspace{2pt}}|@{\hspace{2pt}}c@{\hspace{2pt}}|@{\hspace{2pt}}c@{\hspace{2pt}}}
        \hline
        Bit Width & 24 & 16 & 12 & 8 & 7 & 6 & 5 & 4 & 3 \\
        \hline
        Transition ($\alpha$) Sparsity & 38.05 & 91.47 & 98.18 & 99.51 & 99.65 & 99.75 & 99.83 & 99.90 & 99.94 \\
        \hline
        Emission ($\beta$) Sparsity & 15.28 & 99.19 & 99.87 & 99.96 & 99.97 & 99.98 & 99.99 & 99.99 & 99.99 \\
        \hline
        Initial ($\gamma$) Sparsity & 76.95 & 82.23 & 87.38 & 97.66 & 98.71 & 99.54 & 99.83 & 99.95 & 99.98 \\
        \hline
    \end{tabular}}
    \label{tab:sparsity}
    \vspace{-10pt}
\end{table}

We consider the fixed-point quantization as the candidate for the HMM quantization data format due to its potential to improve the success rate and scores while achieving a promising compression ratio.
First, this approach ensures generalization, as it does not make assumptions about the original distribution. Fixed-point representation uniformly covers numbers between zero and one, making it applicable to all probabilistic models.
Besides, the linear quantization process of fixed-point data, which rounds the numbers to the nearest fixed point, is similar to that of integers. The quantization function is shown below; the scale factor is $2^b$, and the zero point is zero:
\[
    Q_{linear}(p) = clip[round(p * (2^b - 1))] \ /\  2^b.
\]
The rounded results are a feasible and suboptimal solution for the optimization problem, indicating a greedy approach.
Most importantly, the quantization overhead is negligible, and no mapping cookbook needs to be stored. 

However, integer quantization and the fixed-point approach share similar limitations, particularly in terms of generation errors and score loss. 
Besides, the fixed-point linear quantization prunes the weights automatically because numbers will be rounded to zero in the quantization function. As shown in Table~\ref{tab:sparsity}, this auto-pruning by fixed-point linear quantization will exceed the $86\%$ threshold, which is the result that we achieved from ratio-based pruning in Table~\ref{tab:pruning}.
For bit widths below 7, the zero ratios after linear quantization can reach as high as $99\%$.  Therefore, while fixed-point linear quantization offers a feasible solution to the optimization problem, it is not a standalone solution for HMM compression.

\subsection{Norm-Q for Improved Compression}

To prevent information loss during compression and enhance the compression ratio, we propose performing row-wise normalization within weight matrices. That is to say, after fixed-point linear quantization, an extra normalization operation is performed to eliminate errors and further compress the HMM.
Row-wise normalization with fixed-point linear quantization (Norm-Q) is designed specifically for HMM. The normalization process is defined as follows: $\alpha_{ij}\Leftarrow (\alpha_{ij} + \epsilon_j) / \sum_j (\alpha_{ij} + \epsilon_j)$,
where $i$ denotes the row index, $j$ denotes the column index, and $\epsilon$ denotes a small number (e.g., $10^{-12}$).

Row-wise normalization serves to address several critical challenges. 
First, it prevents empty rows and avoids generation mistakes. Values close to zero will be normalized to the same point, controlled by $\epsilon$. This step keeps the sum of each row equal one, which maintains the integrity of the distribution and ensures the correctness of subsequent probability calculations. 
The last column within Table~\ref{tab:pruning} shows that even though $86\%$ is a pruning threshold generating errors,  pruning $86\%$ weights followed by row normalization generates more reasonable results.
However, pruning with the norm method suffers from $18\%$ success rate degradation.
Therefore, Norm-Q is selected based on better generation quality and a high success rate.

Second, the normalization extends the cookbook size while keeping the compression ratio. Each row of the weight matrices is a distribution, and row-wise normalization changes the quantized points of each row separately. The cookbook size far exceeds the number of original fixed-point numbers. A larger cookbook enhances the model's scope and capabilities. The extended cookbook still does not require additional storage.

Our proposed Norm-Q creates a unique path to prevent errors, increase model capacity, and improve compression efficiency, thus ensuring that HMM compression delivers optimal performance with minimal storage overhead. HMM Compression with Norm-Q can achieve low-resolution quantization, with more discussions in Section~\ref{section:experiments_setup}.
Moreover, our experiments show that the weights are automatically pruned by $99\%$ and eventually represented as $b$-bit ($b\le8$) fixed-point numbers. Therefore, the overall compression rate is improved with the Norm-Q method while maintaining the logical constraint success rate and generation quality.

\subsection{Norm-Q Aware Expectation Maximization} \label{section:qaem}

In addition to post-training quantization, we propose to integrate Norm-Q with the HMM training process to facilitate efficient HMM training.
HMM is trained using the expectation maximization (EM) algorithm \cite[]{rabiner1986introduction, moon1996expectation}, which maximizes the expectation of likelihood (LLD) of the distribution. We integrate Norm-Q into the EM process and explore the effect of Norm-Q aware EM training for HMMs. The difference between EM and the training of a neural model is that EM updates the weights by statistical results, and neural networks update by the loss gradient.
The weights are updated during the maximization step, so the weights are quantized after the M step. The weights are quantized every few EM steps, including the last step, as follows:
\noindent
\scalebox{0.9}{
\begin{minipage}{\columnwidth}
\begin{align*}
    \theta^{t+1} = \underset{\theta}{arg\max} \quad E_{Z\sim p(\cdot |X,\theta^t)} [\log p(X,Z|\theta)], \quad \theta\in cookbook^{t+1}.
\end{align*}
\end{minipage}
}

\noindent The quantization interval is a hyperparameter that determines how much training data is consumed before quantization. Different step lengths affect the training efficiency and lead to different accuracies and scores.
The previous linear quantization and normalization methods were performed to solve the maximization problem. Table~\ref{tab:results} presents the Norm-Q aware EM results, and a more detailed discussion is in Section~\ref{section:experiments_normq}.
As an alternative to Norm-Q EM, the results of K-means-aware EM are shown in the last row of Table~\ref{tab:kmeans}. 
The interval equals 20, which means the weights are quantized or clustered by Norm-Q or K-means every 20 EM steps.
Compared to raw K-means, normalized K-means EM increases the success rate by $17\%$ and scores by over 0.6.

\section{Experiments} \label{section:experiments}

\subsection{Setup}
\label{section:experiments_setup}
Our application refers to Ctrl-G \cite[]{lin2019commongen, zhang2025adaptable}, which requires models to generate sentences with concepts or keywords. These keywords must appear in the sentences.
It is challenging because the generative model must satisfy the conditions in the future. The evaluation dataset is composed of 900 references to various concepts.

\begin{figure*}[t]
    \centering
    \includegraphics[width=0.9\linewidth]{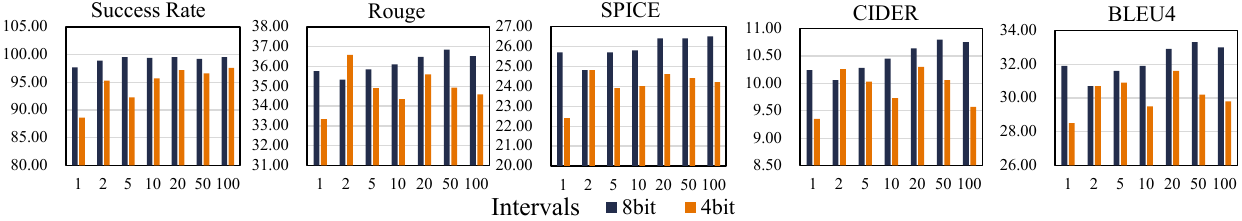}
    \vspace{-8pt}
    \caption{Results (x100\%) of different quantization intervals. }
    \label{fig:qat}
    \vspace{-12pt}
\end{figure*}

In our experiment, the structure of the neuro-symbolic model is as follows: The neural part is an LLM; the symbolic part is an HMM and a deterministic finite automata (DFA). The HMM is distilled from the LLM to assist in generation. The condition is satisfied by adjusting the generating probabilities through the DFA rules and the HMM backward algorithm. The Markov property empowers the ability to predict and change the future generation. The base model is GPT2-large, which is a $774M$ parameter version. The vocabulary size is $50257$.
The beam size of the LLM is set to $128$. The maximum new tokens are set to $32$.
The hidden size of HMM is set to $4096$. Thus, the transition, emission, and initial ($\alpha$, $\beta$, and $\gamma$) matrices have the dimensions of $[4096,4096]$, $[4096,50257]$, and $[1,4096]$, totaling $223M$ parameters.
The dataset for HMM training is sampled from the base model. We sampled $200k$ sentences for training the $223M$ version HMM, and $2k$ sentences are reserved for testing. The training set is divided into $20$ chunks, each containing $10k$ samples.

\subsection{Norm-Q and Norm-Q aware EM Results}
\label{section:experiments_normq}

\begin{table}[ht]
    \centering
    \small
    \caption{Norm-Q and Norm-Q aware EM constraint success rate and score (x$100\%$) results of the basic hmm model}
    \vspace{-5pt}
    \resizebox{\columnwidth}{!}{
    \begin{tabular}{c|c|c|c|c|c|c|c|c}
        \hline
           & FP32 & 8 bit & 7 bit & 6 bit & 5 bit & 4 bit & 3 bit & 2 bit \\
        \hline
        \multicolumn{9}{c}{\cellcolor{gray!30} Norm-Q, HMM hidden size = 4096} \\
        Success Rate & 100 & 99.0 & 99.0 & 97.9 & 97.8 & 97.8 & 96.8 & 91.6 \\
        Rouge & 37.6 & 37.2 & 36.1 & 36.1 & 35.2 & 34.9 & 35.3 & 32.3 \\
        BLEU4 & 35.2 & 34.0 & 32.1 & 32.8 & 31.2 & 30.4 & 30.6 & 25.9 \\
        CIDER & 11.7 & 11.1 & 10.7 & 10.6 & 10.0 & 10.0 & 10.1 & 8.62 \\
        SPICE & 27.0 & 26.7 & 26.0 & 25.6 & 24.8 & 24.7 & 23.9 & 21.8\\
        \hline
        \multicolumn{9}{c}{\cellcolor{gray!30} Norm-Q aware EM, HMM hidden size = 4096} \\
        Success Rate & 100 & 99.0 & 99.0 & 98.1 & 98.2 & 97.2 & 93.1 & 89.3 \\
        Rouge & 37.6 & 36.5 & 36.1 & 36.3 & 36.1 & 35.6 & 34.5 & 33.1 \\
        BLEU4 & 35.2 & 32.8 & 32.2 & 33.1 & 33.2 & 31.6 & 30.7 & 26.7 \\
        CIDER & 11.7 & 10.7 & 10.5 & 10.7 & 10.7 & 10.3 & 9.84 & 8.96 \\
        SPICE & 27.0 & 26.1 & 25.9 & 26.2 & 25.7 & 24.6 & 23.0 & 22.9 \\
        \hline
    \end{tabular}
    }
    \label{tab:results}
    \vspace{-10pt}
\end{table}

There are two types of evaluation metrics. The constraint success rate measures how many outputs satisfy the constraint, i.e., keywords exist. Rouge, BLEU4, CIDER, and SPICE measure the generation quality.
The result of the raw model without pruning or quantization is shown in Table~\ref{tab:int}, denoted as FP32. The scores are lower than the reports from Ctrl-G since the HMM hidden size is smaller in our setup.

Norm-Q enables a higher compression rate with the weight bit number ranging from 8 bits to 2 bits, which outperforms the 14-bit quantization result with layer-wise integer quantization. We select the interval for quantization as 20, and the training epoch is set as 5.  The final rate and score results of Norm-Q and Norm-Q aware EM are shown in Table~\ref{tab:results}.
The results show that the rate loss and each score loss are under $1\%$ for 8-bit quantization. Compared to integer quantization from Table~\ref{tab:int}, the success rate increases by $78\%$, and the scores increase by $7\%$ on average. Compared to K-means from Table~\ref{tab:kmeans}, the success rate increases by $35\%$, and scores increase by $1\%$.
Therefore, Norm-Q improved and balanced the results between the success rate and generation quality.
The ultra-low bit resolution results for Norm-Q are presented. For example, when quantized to 3 bits, the success rate of Norm-Q drops by $3.2\%$, which is much greater than 8-bit K-means and integer methods; the scores drop by $2.9\%$ on average. 

\begin{figure}[H]
    \vspace{-6pt}
    \centering
    \includegraphics[width=0.6\columnwidth]{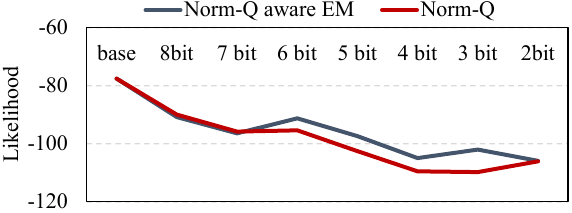}
    \vspace{-6pt}
    \caption{Comparison of likelihoods of Norm-Q aware EM and Norm-Q.}
    \label{fig:quantization}
    \vspace{-8pt}
\end{figure}

We analyze the Norm-Q EM training effect for HMM compression with Norm-Q. In terms of evaluation score,  Norm-Q aware EM has a similar level of performance as Norm-Q, with a difference of less than $1\%$. The reason is that linear quantization is a discontinuous and aggressive operation, leading to unstable convergence. The robustness of the neuro-symbolic model also eliminates the difference.  We further compare the test likelihood between the Norm-Q and Norm-Q aware EM process in Fig.~\ref{fig:quantization}. The overall test likelihood from Norm-Q aware EM outperforms the Norm-Q method, indicating the effectiveness of training with Norm-Q.


We further evaluate the HMM compression ratios of 8-bit and 3-bit, which are  $99.9825\%$ and $99.9992\%$, respectively.
In conclusion, the HMM can be quantized to 8 bits with no success rate loss and quality loss, and 3 bits with an average loss of $3\%$.  Furthermore, Norm-Q can be integrated after standard training or with EM training to provide effective compression. In addition, HMM with Norm-Q achieves an impressive compression rate of over $99.9\%$ while balancing the logical constraints and generation quality.

\subsection{Scalability Analysis}

To validate the scalable performance of our method, the hidden size of the HMM is expanded to 8192 and 16384.
Table~\ref{tab:hmm_scale_results} presents the post-training quantization results.
The constraint success rate of 8 bits remains over $99\%$, and over $96\%$ of 3 bits.
The score loss is under $3.9\%$ and $3.7\%$ on average when hidden size equals 8192 and 16384, respectively.
Overall, Norm-Q demonstrates robust scalability, maintaining the generation quality and accuracy when the HMM size increases, which addresses the deterioration problem.

\begin{table}[t]
    \centering
    \small
    \caption{Success Rate and Score Results (x100\%) of Scaled HMM Models}
    \vspace{-5pt}
    \resizebox{0.75\columnwidth}{!}{
    \begin{tabular}{c|c|c|c|c|c|c}
        \hline
           & FP32 & 12 bit & 8 bit & 6 bit & 4 bit & 3 bit \\
        \hline
        \multicolumn{7}{c}{\cellcolor{gray!30} Norm-Q, HMM hidden size = 8192} \\
        Success Rate & 97.6 & 99.8 & 99.5 & 98.5 & 97.7 & 97.0 \\
        Rouge & 37.7 & 37.4 & 36.9 & 35.5 & 34.2 & 34.2 \\
        BLEU4 & 35.3 & 34.7 & 34.2 & 32.5 & 29.1 & 28.9 \\
        CIDER & 11.5 & 11.4 & 11.1 & 10.3 & 9.6 & 9.6\\
        SPICE & 26.9 & 27.0 & 26.6 & 25.5 & 23.7 & 23.0 \\
        \hline
        \multicolumn{7}{c}{\cellcolor{gray!30} Norm-Q, HMM hidden size = 16384} \\
        Success Rate & 96.5 & 99.7 & 99.5 & 99.3 & 97.9 & 96.4 \\
        Rouge & 37.3 & 37.1 & 36.4 & 36.4 & 34.9 & 33.3\\
        BLEU4 & 34.7 & 34.2 & 33.2 & 33.4 & 31.1 & 29.1 \\
        CIDER & 11.4 & 11.3 & 10.9 & 10.8 & 10.0 & 10.0 \\
        SPICE & 26.7 & 26.8 & 26.4 & 26.0 & 24.8 & 23.0 \\
        \hline
    \end{tabular}
    }
    \label{tab:hmm_scale_results}
    \vspace{-5pt}
\end{table}

\subsection{Quantization Intervals Study for EM Training}

\begin{figure}[t]
    \centering
    \resizebox{0.9\columnwidth}{!}{
    \subfigure[Q-Norm aware EM training LLD]{
        \includegraphics[width=0.5\columnwidth]{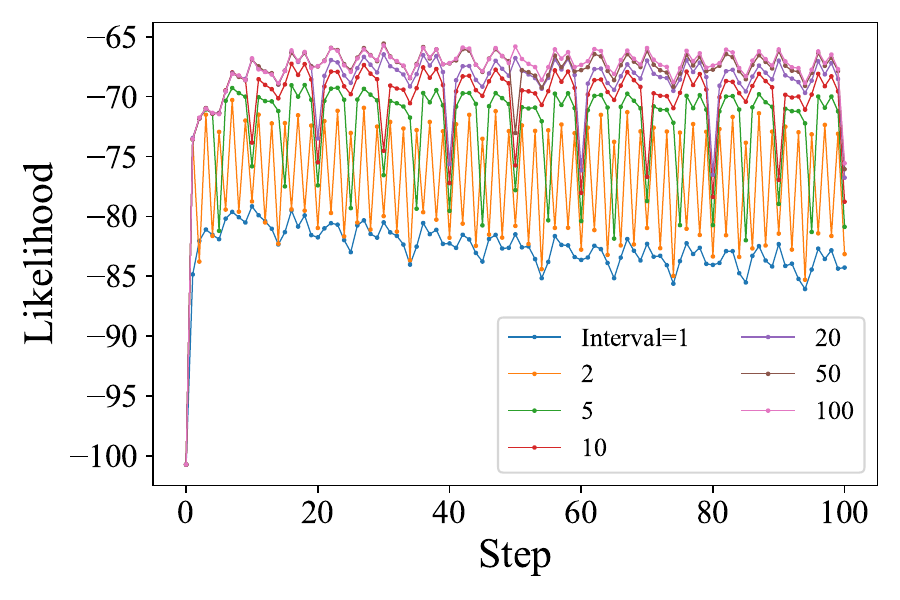}
        \label{fig:train_lld}
    }
    \hspace{-1em}
    \subfigure[Q-Norm aware EM testing LLD]{
        \includegraphics[width=0.5\columnwidth]{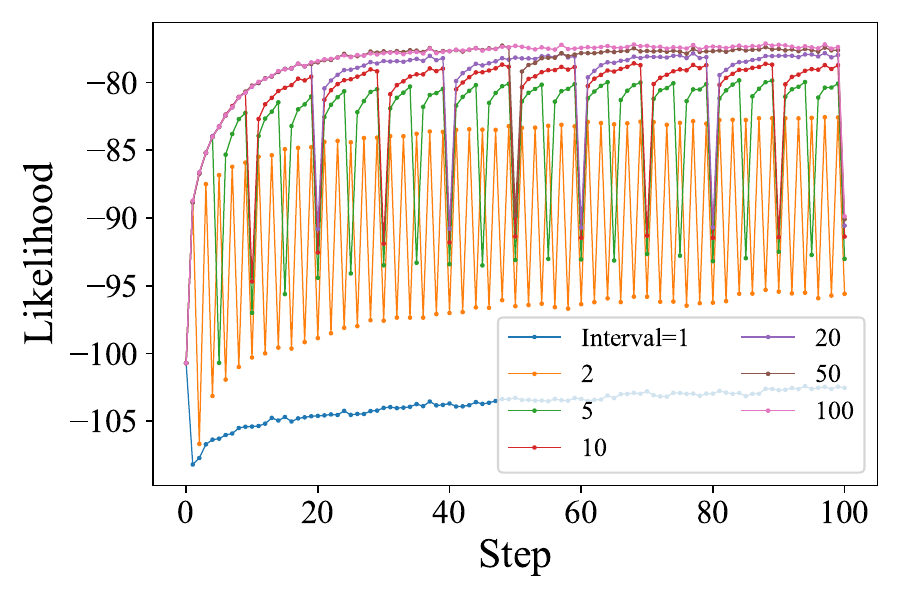}
        \label{fig:test_lld}
    }
    }
    \hspace{-1.5em}
    \resizebox{0.9\columnwidth}{!}{
    \subfigure[LLD results of different intervals]{
        \includegraphics[width=0.5\columnwidth]{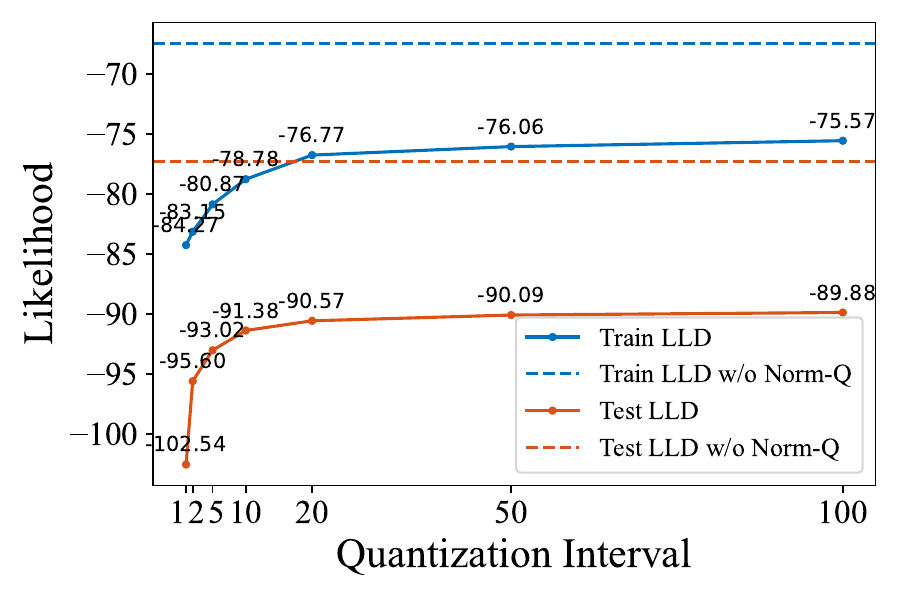}
        \label{fig:final_lld}
    }
    \hspace{-1em}
    \subfigure[LLD of K-means EM]{
        \includegraphics[width=0.5\columnwidth]{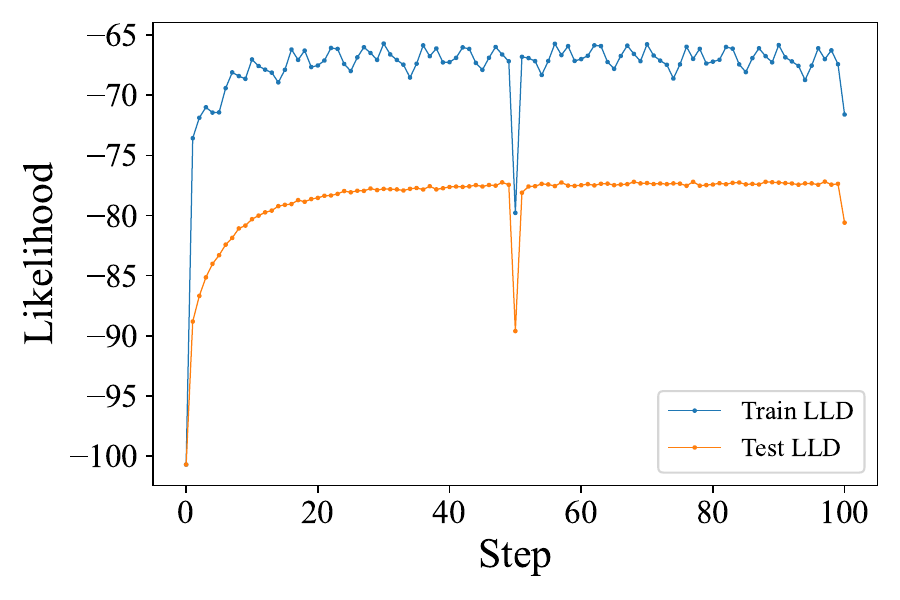}
        \label{fig:kmeans_lld}
    }
    }
    \caption{Likelihood curve during EM. Quantized to 8 bits.}
    \label{fig:curve}
    \vspace{-15pt}
\end{figure}

This section explores the influence of the EM process and the quantization intervals. We experiment with EM of different intervals. The dataset is still 20 chunks, each containing 10k samples. Each EM step consumes one chunk of data. The epoch is 5, which means 100 steps in total.
The likelihood curves during the EM process are shown in Fig.~\ref{fig:curve}, and the final rate and scores are illustrated in Fig.~\ref{fig:qat}. LLD is defined as $E[ \log\ p(X, Z|\theta) ]$ in Section~\ref{section:qaem}, which is our maximization objective.
The up-and-down oscillation of the curve represents the quantization process. Each quantization operation reduces the LLD. The LLD curve oscillates between two bounds, as can be observed from Fig.~\ref{fig:test_lld}. The gap between the upper and lower bounds can be used to measure the quantization loss. Different quantization methods and intervals have different sizes of gaps. We integrate the EM training method with K-means to validate the effectiveness of training in Table~\ref{tab:kmeans}. Fig.~\ref{fig:kmeans_lld} demonstrates the LLD curve of the K-means method, which converges similarly.
The LLD converges around step 30, meaning that the model is stable after training with 30 chunks of data. The converging point relates to the hidden size.

The final LLD after 100 steps is shown in Fig.~\ref{fig:final_lld}. As the interval increases, the final LLD also increases. There is a converging threshold for the intervals, which is 20 in this case. When the interval exceeds this threshold, the final LLD stabilizes. In our previous experiments, we chose an interval of 20 because larger intervals lack sufficient training and fine-tuning, potentially leading to a larger quantization gap. On the other hand, smaller intervals result in poorer LLD and slower convergence. Fig.~\ref{fig:qat} explores the design space of quantization bits (4 and 8), and intervals (1, 2, 5, 20, 50, and 100) with the consideration of the success rate and scores. Our experiments demonstrate that an interval of 20 performs best at 4 bits, while an interval of 50 outperforms others at 8 bits.

\section{Conclusion and Future Work}

This paper works on compressing hidden Markov models, or large probabilistic models in general, in neuro-symbolic applications to reduce memory and bandwidth pressure for the system. We experiment with several techniques to quantize the model, including pruning, layer-wise quantization, clustering, integer method, and our Norm-Q method. We successfully quantize various sizes of HMMs to 8 bits without loss and 3 bits with acceptable loss, demonstrating effectiveness and scalability. The models are compressed over $99\%$.
For future work, co-optimization of the neural and symbolic parts will investigate the interactions and synergy between two parts. 
Besides, compression and quantization like Norm-Q require dedicated hardware support. 

\section*{Acknowledgment}

This work was supported by the Research Computing Group and the High Performance Computing systems Afton and Rivanna at the University of Virginia, which provides necessary computing resources.

\bibliographystyle{unsrt}
\bibliography{refs}

\end{document}